\documentclass[letterpaper, 10 pt, conference]{Settings/ieeeconf}  

\IEEEoverridecommandlockouts                              

\usepackage{footnote}
\usepackage{amsmath} 
\usepackage{amssymb}  
\usepackage{booktabs}
\usepackage{color}
\usepackage{xcolor}
\usepackage{multirow}

\usepackage[pdftex]{graphicx}
\usepackage{siunitx}
\usepackage{cite}

\usepackage{hyperref}
\usepackage{subcaption}
\usepackage{glossaries}

\newacronym{HRI}{HRI}{human-robot interaction}
\newacronym{EBC}{EBC}{Expectation-Based Control}
\newacronym{MPC}{MPC}{Model Predictive Control}
\newacronym[longplural={degrees of freedom},shortplural={DoFs}]{DoF}{DoF}{degree of freedom}
\newacronym{SMU}{SMU}{Safe Motion Unit}
\newacronym{SSM}{SSM}{speed and separation monitoring}
\newacronym{PFL}{PFL}{power and force limiting}
\newacronym[longplural={inertial measurement units},shortplural={IMUs}]{IMU}{IMU}{inertial measurement unit}
\newacronym[longplural={evasive motions},shortplural={EMs}]{EM}{EM}{evasive motion}
\newacronym[longplural={Expectable Motion Units},shortplural={EMUs}]{EMU}{EMU}{Expectable Motion Unit}
\newacronym{FE}{FE}{Franka Emika}
\newacronym{IMO}{IMO}{human involuntary motion occurrence}
\newacronym{TUI}{TUI}{Technology Usage Inventory}
\newacronym{PS}{PS}{perceived safety}
\newacronym{GS}{GS}{Godspeed}

\usepackage{fancyhdr}
\definecolor{excelblue}{HTML}{4472C4} 
\definecolor{NewBlue}{HTML}{4472C4} 
\definecolor{NewOrange}{HTML}{ED7D31} 
\definecolor{NewGreen}{HTML}{00B050} 

\usepackage{stackengine}
\newcolumntype{P}[1]{>{\centering\arraybackslash}p{#1}}

\input{Settings/abkuerzung.sty}
\graphicspath{{graphics/}}
\DeclareGraphicsExtensions{.pdf,.jpeg,.png}

\renewcommand{\baselinestretch}{0.94}

\def\mytitle{Towards Safe Robot Use with Edged or Pointed Objects: A Surrogate Study Assembling a Human Hand Injury Protection Database}
\title{\LARGE \bf \mytitle}

\def\myauthor{Robin Jeanne Kirschner$^{1}$, Carina M. Micheler$^{2,3}$, Yangcan Zhou$^{1}$, Sebastian Siegner$^{1}$, Mazin Hamad$^{1}$, \\Claudio Glowalla$^{2,4}$, Jan Neumann$^{5}$, Nader Rajaei$^{1}$, Rainer Burgkart$^{2}$ and Sami Haddadin$^{1}$}
\def\mythanks{$^1$ Chair for Robotics and Systems Intelligence, Munich Institute of Robotics and Machine Intelligence, Technical University of Munich, 80992 Munich, Germany\newline 
$^2$ Department of Orthopaedics and Sports Orthopaedics, Klinikum rechts der Isar, TUM School of Medicine, Technical University of Munich, 81675 Munich, Germany\newline
$^3$ Institute for Machine Tools and Industrial Management, TUM School of Engineering and Design, Technical University of Munich, 85748 Garching near Munich, Germany\newline
$^4$  Endoprothetikzentrum der Maximalversorgung, BG Hospital Murnau, 82418 Murnau, Germany\newline
$^5$ Department of Diagnostic and Interventional Radiology, Klinikum rechts der Isar, TUM School of Medicine, Technical University of Munich, 81675 Munich, Germany\newline
Corresponding author:
{\href{mailto:robin-jeanne.kirschner@tum.de}{\tt\small robin-jeanne.kirschner@tum.de}}
}

\author{\myauthor
\thanks{}
\thanks{\mythanks}%
}

\hypersetup{ 
pdftitle={\mytitle},
pdfauthor={\myauthor},
}

\DeclareUnicodeCharacter{2009}{\,}

\begin{document}

\maketitle
\thispagestyle{fancy} 
\pagestyle{fancy} 

\begin{abstract}


The use of pointed or edged tools or objects is one of the most challenging aspects of today's application of physical human-robot interaction (pHRI). One reason for this is that the severity of harm caused by such edged or pointed impactors is less well studied than for blunt impactors. Consequently, the standards specify well-reasoned force and pressure thresholds for blunt impactors and advise avoiding any edges and corners in contacts.  
Nevertheless, pointed or edged impactor geometries cannot be completely ruled out in real pHRI applications. For example, to allow edged or pointed tools such as screwdrivers near human operators, the knowledge of injury severity needs to be extended so that robot integrators can perform well-reasoned, time-efficient risk assessments. In this paper, we provide the initial datasets on injury prevention for the human hand based on drop tests with surrogates for the human hand, namely pig claws and chicken drumsticks. We then demonstrate the ease and efficiency of robot use using the dataset for contact on two examples. Finally, our experiments provide a set of injuries that may also be expected for human subjects under certain robot mass-velocity constellations in collisions. To extend this work, testing on human samples and a collaborative effort from research institutes worldwide is needed to create a comprehensive human injury avoidance database for any pHRI scenario and thus for safe pHRI applications including edged and pointed geometries.

\end{abstract}

\glsresetall 
\section{INTRODUCTION}
 
 \begin{figure}[th]
	
		\includegraphics[width=1\linewidth]{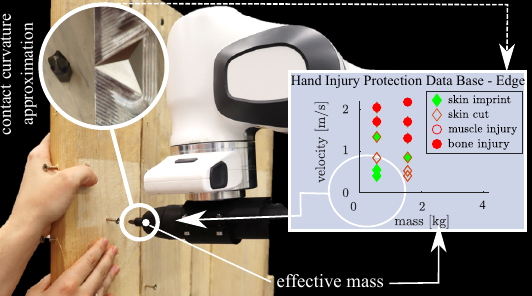}
	\caption{Using the generalized hand injury protection database for a constrained contact with a sharp edge can allow a reasonable risk assessment for contacts with a Phillips head, provide permissible mass-velocity combinations for safe impact scenarios, and allow deployment of safe velocity-scaling based on the computable effective robot mass.}
	\label{fig:specimen}
	
	\vspace{-7mm}
	\label{fig:intro}
\end{figure}

Safety in human-robot interaction is one of the major challenges that need to be addressed for successful robot deployment \cite{Haddadin.2012, Lachner_2021}.
Especially with the trend of flexible task execution in industrial assembly lines, collaborative human robot application and contact-rich manipulation are becoming increasingly important. Tasks that are being automated using robots are for instance, (un)screwdriving \cite{Villa_2022,Hjorth_2023}, collaborative manufacturing \cite{Tassi_2022}, or handling of electronics parts \cite{casalino_2018}. Enabling, understanding and controlling contacts is a fundamental prerequisite for this, which is made possible, e.g., by impedance control/uniform impedance \cite{Karacan_2023, Hjorth_2023}. 

Nevertheless, errors may still occur, e.g. slipping off the surface or misplacing the tool or objects.
In these cases proximity between human and robot can result in hazardous scenarios like poking the hand with pointed tool tips, cutting the skin with edges of grippers or objects, or even bone injuries due to deep cuts with sheet objects. To enable such applications nevertheless, requires an appropriate risk assessment and risk mitigation measures \cite{iso12100}.  The categorization of possible injury events according to their severity is fundamental to this risk assessment. The currently available standards provide a dataset for permissible contact forces and pressures based on experimentally evaluated human onset of pain in blunt contacts \cite{iso15066}. The compliance to these forces and pressures must be validated for the robotic application. This means that impact forces and pressures must be measured in an iterative process requiring costly test devices and additional test setups \cite{Lachner_2021, Fischer_2022}, which leads to long downtimes in the production line every time parameters in the application are changed \cite{Kirschner_2022_iso}. 
This makes the cost- and time-efficient use of robots in unstructured or dynamically changing environments more difficult and ultimately limits the potential for the next generation of flexible automation.


\begin{table*}[t]
 
	\begin{minipage}[t]{\linewidth}
	
    \centering
    \caption{Overview of studies reporting severity of harm for human hand available for risk assessment based on \cite{iso12100}}
    \begin{tabular}{p{6mm}p{18mm}p{7mm}P{6mm}P{6mm}P{6mm}P{6mm}P{9mm}P{9mm}P{6mm}P{6mm}P{6mm}P{6mm}P{9mm}P{9mm}}
    \toprule
        \multirow{3}{*}{\textbf{Cat.}} & \multirow{3}{*}{\textbf{Injury}} & \multirow{3}{*}{\textbf{Studies}} & 
        \multicolumn{6}{c}{\textbf{Subjects}} & \multicolumn{4}{c}{\textbf{Injury parameters}}  & \multicolumn{2}{c}{\textbf{Impactor types}}  \\ 
        & & & \multicolumn{2}{c}{human} & \multicolumn{2}{c}{surrogate} & \multirow{2}{*}{synthetic} & \multirow{2}{*}{simulation} & \multirow{2}{*}{$F$} & \multirow{2}{*}{$p$} & \multirow{2}{*}{$[m,v]$} & \multirow{2}{*}{$E$}  & \multirow{2}{*}{edged} & \multirow{2}{*}{blunt}  \\
         & & &  iv & ev & iv & ev & \\
         
        \midrule
        
         \multirow{7}{*}{\textbf{S0}} &  \multirow{7}{*}{Onset of pain} & \cite{iso15066} & 
         \checkmark & - & - & - & - & - & \checkmark & \checkmark & - & - & - & \checkmark \\
         
         & & \cite{Behrens_2014} & 
         \checkmark & - & - & - & - & - & \checkmark & \checkmark & \checkmark & \checkmark & - & \checkmark \\
         
         & & \cite{Behrens_2022} & 
         \checkmark & - & - & - & - & - & \checkmark & \checkmark & - & - & - & \checkmark \\
         
         & & \cite{Yamada_1996} & 
         \checkmark & - & - & - & - & - & \checkmark & - & - & - & - & \checkmark \\
         
         & & \cite{Asad_2023} & 
         - & - & - & - & - & \checkmark & \checkmark & \checkmark &  (\checkmark) & \checkmark & - & \checkmark \\
         
         & & \cite{Hohendorff201339} & 
         \checkmark & - & - & - & - & - & \checkmark & - & - & - & (\checkmark) & - \\
         
         & & \cite{Povse_2010}\footnote{Data for human lower arm\label{underarm}} &
         \checkmark & - & - & -& - & - & \checkmark & - & - & \checkmark & (\checkmark) & \checkmark \\
         
         & &  \cite{Hohendorff20112158} 
         & - & \checkmark & - & - & - & - & \checkmark & - & - & - & (\checkmark) & 
         - \\

         & & \cite{Mewes_2003} & \checkmark & - & - & - & \checkmark & - & \checkmark & - & - & - & (\checkmark) & \checkmark\\
         
         \midrule
         \multirow{1}{*}{\textbf{S1}} & \multirow{1}{*}{skin contusion} & \cite{Behrens2023} 
         & \checkmark & - & - & - & -& - & \checkmark & \checkmark & (\checkmark) & \checkmark & - & \checkmark \\
         
         
        \midrule
         \multirow{2}{*}{\textbf{S2}} & \multirow{2}{*}{bone fracture}  &  \cite{Hohendorff201339} &
         - & \checkmark & - & -& - & - & \checkmark & - & - & - & (\checkmark) & - \\
         & &  \cite{Kent_2008} &
         - & \checkmark & - & - & - & - & \checkmark & - & - & - & (\checkmark)  & -\\

         \bottomrule
    \end{tabular}
    \vspace{-3mm}
    \label{tab:injury_severity}
    \end{minipage}
    \vspace{-7mm}
\end{table*}

Understanding the dynamics between human injury and collision scenarios in physical human-robot interaction (pHRI) based on controllable and observable parameters such as energy-based parameters like effective mass, velocity, and contact curvature is a fundamental step to enable efficient and well reasoned risk assessment \cite{Haddadin.2012, Lachner_2021} as depicted in Fig. \ref{fig:intro}. This injury knowledge can best be generated by human injury analysis which requires consideration of all human body parts, and relevant medical factors such as age. Hence, a large collection of datasets resulting from a multitude of experiments using adequate surrogates or human tests is required \cite{Behrens2023, Haddadin.2012}. Historically, injury testing was performed for deadly car crashes \cite{Ndiaye_2009,Haynes_1961} and severe clamping situations \cite{Hohendorff20112158}. 
 Ensuring safety in pHRI, nevertheless, requires consideration of injuries occurring in contact with lower effective masses and contact velocities but varying contact geometries \cite{Haddadin.2012}.
 Especially, the human hand is typically exposed to high injury potential in pHRI \cite{Asad_2023}. In order to create a human injury protection database that can be efficiently used for risk assessment, in this paper we first discuss the available datasets on human hand injury protection in terms of the severity of harm suggested by industry standards. Building on this, we propose studies to assess injury cases caused by edged and pointed impact devices. To this end, we present two surrogates for the analysis of human hand injuries based on anatomical similarities and analyze their recurrent impact injury patterns. We then develop testbed, procedures, and evaluation methods for a generalizable injury analysis based on curvature, mass, and velocity of impact. Using 351 experiments with pig dew claws and 117 experiments with chicken drumsticks as surrogates for the human hand, we derive, to the best of the authors' knowledge, the first database for protecting the human hand from injury in edged and pointed impact geometries using tetrahedral, wedge-shaped, and sheet-shaped impactors in constrained contact scenarios. Finally, we demonstrate the application of the injury protection database using the example of screwdriving and handling of electronic components and discuss future developments.
 This paper is structured as follows. Sec. \ref{sec:State of the Art} provides an overview of the state of the art in human hand injury analysis.  Then, Sec. \ref{sec:methodology} motivates and describes the conducted injury experiments and validation study. Following Sec. \ref{sec:results} and Sec. \ref{sec:Discussion} list and discuss the results and Sec. \ref{sec:Conclusion} concludes the paper.

\section{Available Human Hand Injury Data}\label{sec:State of the Art}

Enabling risk assessment requires estimation of the severity of harm resulting in a impact \cite{iso12100}. Consequently, for pHRI deployment, we require datasets relating human injury occurrence to robot parameters \cite{Haddadin.2012}. For this, three categories of fundamental information are required to relate the available injury datasets to the concrete automation application at hand: 
\begin{itemize}
    \item[a)] Subject type (including body location),
    \item[b)] Measured injury-inducing parameters, and
    \item[c)] Considered impactor geometry.
\end{itemize}

Subjects for injury experiments are typically in-vivo (iv) or ex-vivo (ev) human specimen \cite{Behrens_2014} or animal surrogates \cite{Haddadin.2012}, synthetic specimen like dummies \cite{haid_2023}, or even simulations \cite{Asad_2023}. General injury parameters reported in literature are forces $F$ \cite{iso15066}, pressure $p$ \cite{iso15066}, mass and velocity pairs $[m,v]$ \cite{Haddadin.2012}, and energy-based metrics $E$ such as impact energy density\cite{Povse_2010}, total transferred energy \cite{Sugiura_2019}, or impulse\cite{Hohendorff20112158}. Lastly, the impactor geometries that encodes the impact curvature are variable, but can generally be distinguished into blunt or edged impactor geometries. Here, we classify boned but not rounded impactors as edged \cite{Povse_2010,Haddadin.2012} and flat, rounded impactors with contact surfaces of $\approx$ \SI{1}{cm^2} as blunt \cite{Behrens_2014}. 
Table \ref{tab:injury_severity} lists available human hand injury biomechanics datasets to the best of the authors' knowledge, in line with our previous literature reviews \cite{Hamad_2021_1,Hamad_2021_2}\footnote{(\checkmark) marks impactors with $A <$ \SI{1}{cm^2} but rounded edges  for \emph{edged} and incomplete data for \emph{[m,v]}.}.

 It is evident that especially impacts with edged impactors are not well researched for adequate risk assessments, although fully avoiding edged contacts does not seem feasible considering pHRI applications in industrial scenarios. Moreover, the directly controllable/computable robot parameters (namely velocity and effective mass) that cause injuries are hardly reported, which complicates the integration process into robot control. This is also reflected by the state of the art standardization, which provides force and pressure thresholds based on contact geometries $\geq$ \SI{1}{cm^2} and demands force and pressure analysis in contact by experimental risk analysis \cite{iso_10218-2}. To systematically address this gap and enable efficient risk assessment for any robotic application (including edged contact geometries), we derive an initial datasets for ex-vivo human hand surrogates.

\section{Methodology} \label{sec:methodology}

	
	


This section describes the test conditions, the evaluation procedures and the test setups used to derive human injury protection datasets. It also describes the validation experiment to demonstrate the application of the generated injury protection datasets for risk reduction.



\begin{figure}[t]
		\includegraphics[width=1\linewidth]{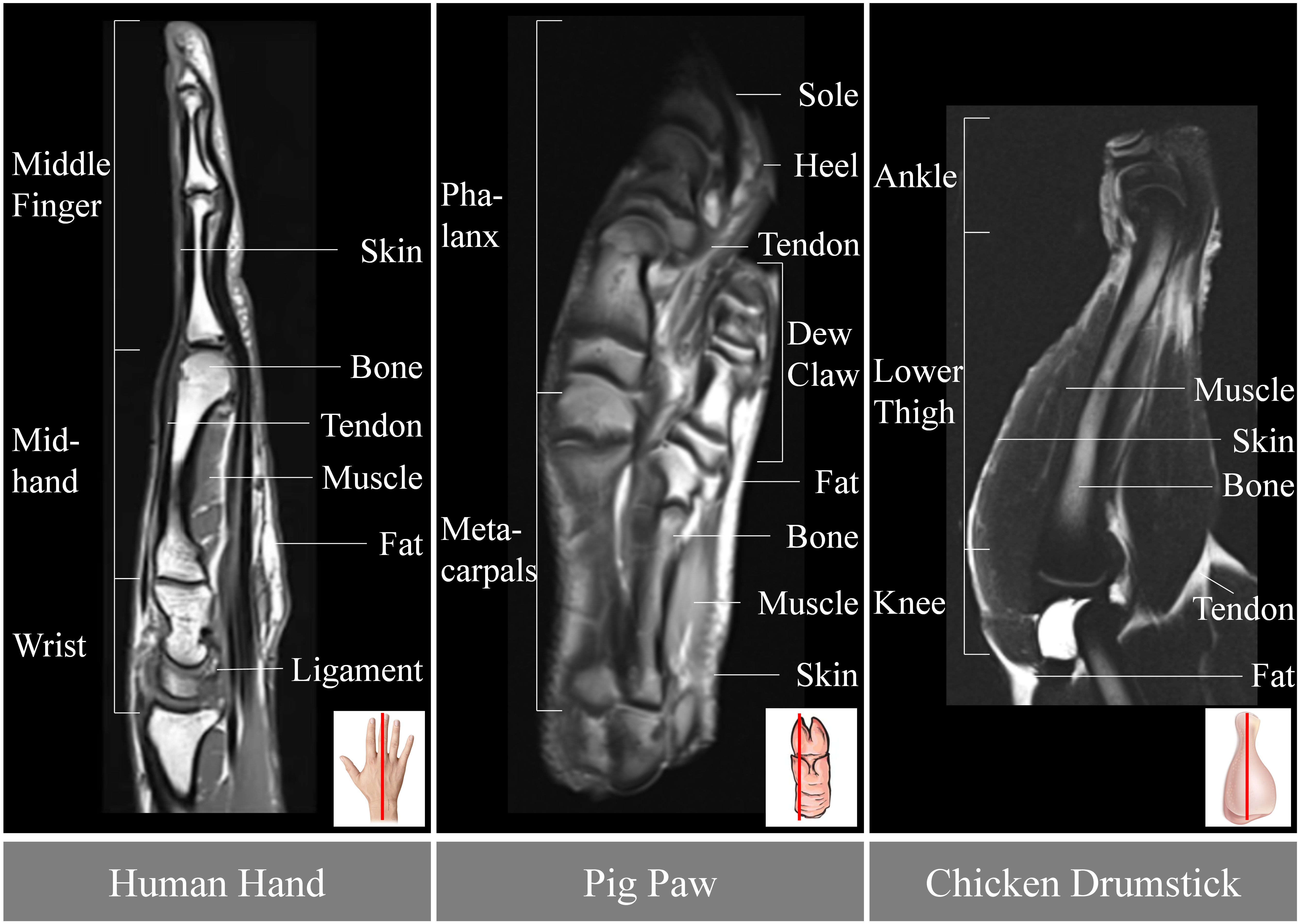}
	\caption{MRI scan and cross-section of the different experiment specimen as compared to the human hand \cite{Micheau_2015}, key tissue types indicated for different specimen.}
	\label{fig:MRI}
	\vspace{-6mm}
\end{figure}

\subsection{Experimental Subject Selection}
This study focuses on hand injuries, as these are the parts of the body that interact most with collaborative robots. Various types of surrogates are suitable as experimental subjects. Synthetic models are ethically unobjectionable. However, their ability to replicate biomechanical properties is limited. Animal and cadaver specimens come closer to the biomechanical conditions of living humans and are preferred in this experimental setup. In the literature, porcine specimens are the most commonly used animal specimens \cite{ankersen1999puncture,Shergold2006,Germscheid2011,Marchant-Forde2018,Sugiura_2019,Haddadin.2012}, as they approximate the biomechanical properties of the human. 
 The front feet of a pig with the corresponding soft tissues can be used to mimic the human hand, and the pig toes mimic the general structure of human fingers well, as shown in Fig. \ref{fig:MRI} is shown. The metacarpal and proximal phalanx of the pig claw best resemble human fingers, as they are thinner than their counterparts. One limitation is that the skin of pigs is thermally treated during slaughter, which changes its structural properties and makes the skin tougher. On this basis, chicken thighs (Gallus domesticus) are chosen as a second surrogate to provide a reference in particular for the range of skin lesions that can occur in real human specimens, although the chicken's anatomical similarity to humans is limited, especially due to the bird's hollow bones.
The average thickness of chicken skin (epidermis + dermis) can range from \SI{0.6}{} to \SI{1.5}{mm}, depending on age, feeding and skin area \cite{Oliveira2022,V.R.2020}. In contrast, the thickness of porcine skin varies between \SI{1}{} and \SI{6}{mm} \cite{Shergold2006,Summerfield2015}. The human skin (epidermis + dermis) on the back of the hand in comparison is about \SI{2.3}{mm} thick \cite{Oltulu2018}.

In this study, we use porcine specimens (dew claws), as previously applied in the above listed literature and compare them with chicken drumsticks. 


	
	



\begin{figure}[t]
\centering
		\includegraphics[width=0.9\linewidth]{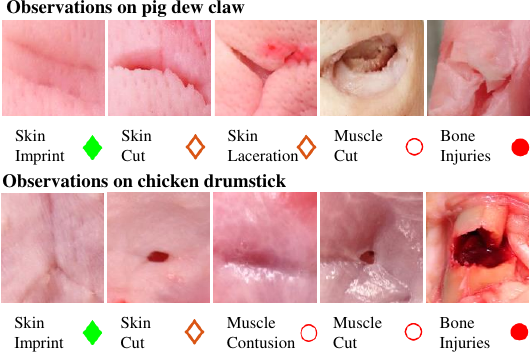}
		
	\caption{Observable features representing injury types occurring upon impacting the pig dew claw and chicken drumstick.}
	\label{fig:chicken_injuries}
	\vspace{-2mm}
\end{figure}


		


		

\subsection{Injury Evaluation}
Classification patterns for fractures associated with soft tissue injuries include the so-called AO classification established by the Association of the Study of Internal Fixation \cite{ruedi2000principles}, the Gustilo-Anderson classification \cite{gustilo} and the Oestern and Tscherne classification \cite{Oestern1984}.
Based on these classifications, in a preliminary study, we noted the observable injuries made in impacts with the test specimens which were labelled by a medical professional as exemplary depicted in Fig. \ref{fig:chicken_injuries}. For the pig dew claws we observed skin imprints, skin injuries (including cuts and lacerations), muscle injuries, and bone injuries (including visible imprints and fractures) as depicted on the top of Fig. \ref{fig:chicken_injuries}. For the chicken drumsticks we also observed skin imprints, skin cuts, muscle injuries, and bone injuries as presented on the bottom of Fig. \ref{fig:chicken_injuries}. For visual analysis of the skin injuries we first analyse the specimen epidermis superficially while for the bone and fibre injuries we open the specimen to the layer of interest. The different injuries types are further referred to using the presented marker in Fig. \ref{fig:chicken_injuries}.

\subsection{Impact Experiments}



\subsubsection{Experimental Setup Design}

\begin{table}[tpb]
    \centering
    \caption{Experimental settings}
    \vspace{-0.1cm}
    \begin{tabular}{p{16mm}p{10mm}p{27mm}p{15mm}}
        \toprule
        Parameter & & Value  & Motivated by   \\ 
        \midrule
         effective mass & & \SI{0.5}{}-\SI{8}{kg} &  \cite{Kirschner_2021_Notion}\\
         velocities & & \SI{0.20}{}-\SI{2.0}{m/s} &  \cite{iso_10218-2}, \cite{franka_FR3_da}\\
         \multirow{3}{*}{\parbox{16mm}{impact geometries}} & wedge & prism \ang{90}, boned &  \cite{Haddadin.2012}\\
          & edge  & tetrahedron \ang{90}, boned &  \cite{Haddadin.2012}\\
           & sheet  & width \SI{1.5}{mm}, boned &  \cite{casalino_2018}\\
        \bottomrule
    \end{tabular}
    \label{tab:Experimental parameters}
    \vspace{-0.5cm}
\end{table}

\begin{figure}[t]
\centering
\includegraphics[width=0.9\linewidth]{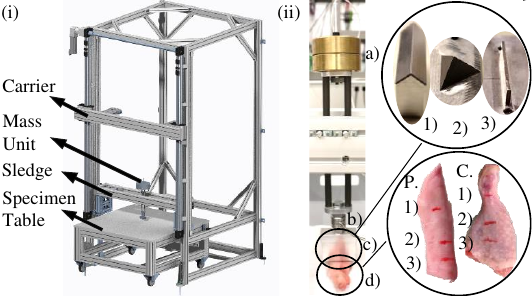}
\caption{Impact drop testing and subject placement: Drop test device (i), mass unit with impactor and specimen (ii) and the applied impactors and prepared specimen, pig (P) and chicken (C) with three distinct impact locations (1-3 from distal to proximal).}
\vspace{-0.3cm}
\label{fig:DT}
\end{figure}

Drop testing is a common procedure for material testing, which is lately also applied for injury analysis in sports \cite{haid_2023} and robotics  \cite{Haddadin.2012}, motivating the design of our test bed. The developed drop test device is depicted in Fig. \ref{fig:DT} (i). Its control and data acquisition runs via a compactRio system from company National Instruments and samples all data at \SI{2000}{Hz}. The linear guided sledge is lifted up by electromagnets attached to a carrier system on linear units as presented in Fig. \ref{fig:DT}. 
 On the sledge, a falling mass unit is integrated (c.f. Fig. \ref{fig:DT} (ii)) that yields desired masses ranging from \SI{0.6}{kg} up to maximum \SI{40}{kg}. The additional mass is added on the mass unit as depicted in Fig. \ref{fig:DT} a). At the end of the falling mass unit the impactor is mounted, see Fig. \ref{fig:DT} c). In this study, we chose the following impactors (check Table \ref{tab:Experimental parameters} for details) 
 \begin{itemize}
     \item \textit{Wedge (W)}; Fig. \ref{fig:DT} 1),
     \item \textit{Edge (E)}; Fig. \ref{fig:DT} 2), and
     \item \textit{Sheet (S)}; Fig. \ref{fig:DT} 3).
 \end{itemize} 
 
These impactors are made from aluminium alloy EN AW-7075 with hardness \SI{150}{HB}. The impact velocity is obtained by a calibration curve relating the relative height to the impact velocity measured using a DAkkS-calibrated precision light barrier (Precision lightbarrier 203.10 from company Hentschel \cite{Hentschel_datasheet}). The calibration is obtained by a measurement routine where the light barrier is positioned at the point of contact and a calibration object crosses the laser beam right before collision from different relative drop heights. An uni-axial piezo-electric force sensor (Type 9331C from Kistler \cite{force_datasheet}), Fig. \ref{fig:DT} b), with amplifier LabAmp 5165A \cite{labamp_datasheet} running at \SI{10}{kHz} is integrated to measure the contact force resulting on the probes under the impactor, see Fig. \ref{fig:DT} d). For safe placing of the probes under the drop test by the operator, a table with linear unit, that can be moved out of the testing area, was integrated.







\subsubsection{Experimental Protocol}

\begin{figure}[t]
\centering
\includegraphics[width=1\linewidth]{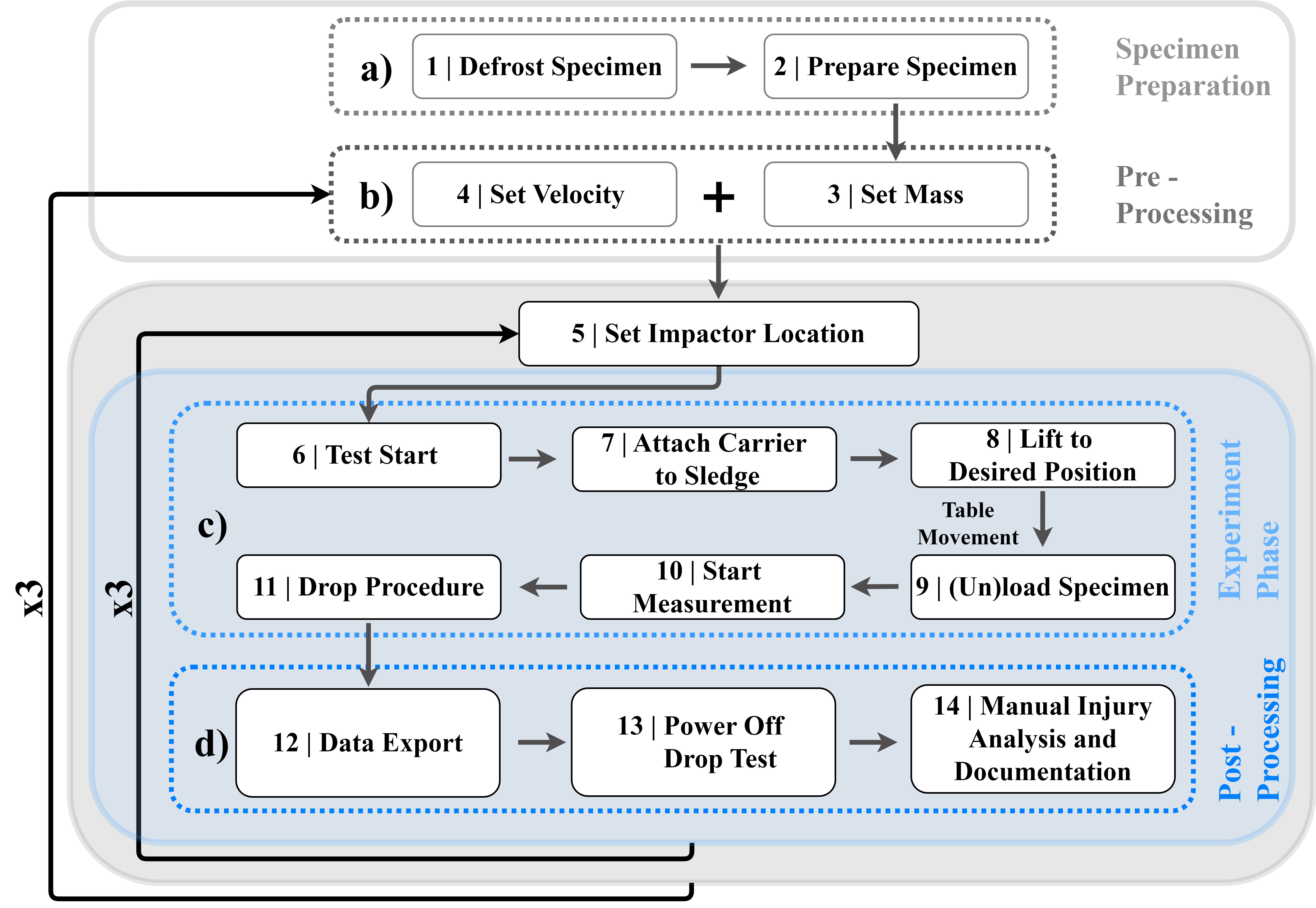}
\caption{Experimental Flowchart.}
\vspace{-0.5cm}
\label{fig:protocol_flowchart}
\end{figure}

The overall experimental protocol is depicted in Fig. \ref{fig:protocol_flowchart}. In the following, we detail the experimental steps.

\textbf{Preparation procedure:}
As summarized in Fig. \ref{fig:protocol_flowchart} a) and b), the preparation procedure first requires the frozen samples used on a test day ($\approx$ 4 pig feet or 8 chicken legs) to be thawed at room temperature for approx. 10 hours. The samples are then prepared by dissecting the individual pig dew claws and chicken legs. The specimen are examined for existing injuries, which are marked, and finally the specimen appearance is documented photographically. Then, the impact locations are indicated by skin markers, excluding the dew claw joints for the pork specimen. Then, the desired dropping height based on the desired velocity and impact mass are adjusted. Additionally, the positioning of the specimen is adjusted, such that the desired impact location is reached. Currently unused samples were stored with moisture to prevent structural skin changes.

\textbf{Experimental Phase:}
The drop test device is used to cause the impacts following the procedure depicted in Fig. \ref{fig:protocol_flowchart} c) and d). First, the device is started. Then, the carrier moves in contact to the sledge and pulls the the sledge up to the defined height. To load the specimen, the table moves out. The specimen placement is adjusted by the operator. Afterwards, the operator triggers the test from the PC and the table automatically moves back to its initial position. After the table moved in place, all measurement devices are activated. Lastly, the magnets release the sledge, which falls and is then caught by a damper unit. Consequently, only the detached mass unit strikes the sample at the desired location and at the speed obtained from the drop height. This process is repeated at all three defined impact locations of each sample and with three freshly prepared samples. Finally, the procedure is repeated with the next mass-velocity configuration.

\textbf{Postprocessing:}
 Once the device turned off, the examiner fetches the probe for a visual analysis of the skin. Then, the specimen is opened and bone or tissue injuries are observed. All findings are documented photographically and video recordings are stored for later reference.
 The drop testing procedure causes consecutively a dynamic (transient) collision force of short duration ($<$\SI{0.5}{s}) followed by the impactor resting on the specimen as quasi-static load \cite{iso15066}. As the resting load from our dropping weight causes quasi-static forces of $\sim$ \SI{6}{N} - \SI{36}{N}, while the impacting forces are six to 30-times higher, we assume the observed injuries to result from the initial, dynamic impact phase.

\subsection{Validation study}
 \begin{figure}[t]
	
		\includegraphics[width=1\linewidth]{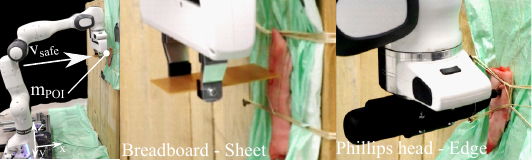}
	\caption{Validation scenarios screw driving and breadboard handling with constrained contact between the surrogate for the human hand and the Phillips head/bread board to the wall.}
	\label{fig:val_setup}
	
	\vspace{-5mm}
\end{figure}
To substantiate the applicability of the established database for safe robot deployment, we show the according risk reduction procedure based on our generated datasets for edged and pointed impactors using a screw driving \cite{Hjorth_2023} and electronics (breadboard) handling \cite{casalino_2018} application. The Phillips head of the screwdriver, can conservatively be simplified with our edged impactor geometry\footnote{Note that the \emph{chamfered} edges of the Phillips head are expected to cause less shear than the only \emph{slightly boned} edges of the pointed impactor E.}. The breadboard is of same thickness the sheet impactor and both are only boned. To enable efficient tool-usage with such geometries, the first requirement is that the effective mass at the constrained impact location should be as low as possible. For this, we obtain the robot joint configurations allowing low effective masses at the point of interest $m_\mathrm{POI}$ by simulation following the effective mass evaluation in \cite{Hamad_2023,Hamad_2019, Khatib_1995}. The robot configuration $q = [0.013,-0.305,-0.069,-1.90,-0.02,1.59,0.78]^\tp$ was found ideal to conduct both example tasks and provide a low effective mass of $m_\mathrm{POI} =$ \SI{1.50}{kg} and $m_\mathrm{POI} =$ \SI{1.92}{kg} during motion in $x$-axis direction of the base frame with the Cartesian unit direction being $\vu = [1\, 0\, 0]$ for the screw driver and breadboard setup, respectively. Based on the previously established hand injury protection database, one can approximate the maximum permissible velocities by, e.g., considering skin imprints as only acceptable observations.




As a validation for each setup pig dew claws are used as human finger surrogates as depicted in Fig. \ref{fig:val_setup}. The robot was programmed such that a relative Cartesian motion of \SI{30}{cm} is performed by moving the end-effector towards the wooden wall at safe velocity according to the previously generated datasets. During motion the tip of the screwdriver or edge of the breadboard respectively collides with the pig specimen approximately after \SI{15}{cm}. Directly after contact the skin injury is examined and documented. Then, the impact location is changed aiming for at least six contact points\footnote{Joint spaces and regions of visible previous injury were excluded.}. This procedure is conducted for three different pig dew claws, at the proximal phalanx and metacarpal of the dew claw. After all impacts were successfully established, the specimen is dismounted from the test location and the superficial skin (epidermis) injuries are examined visually. Then, the specimen is placed in the fridge for an hour and remaining skin injuries are reported. Lastly, deeper skin (dermis), muscle, and bone injuries were examined by dissection. 

\begin{figure*}[t]
\centering
\includegraphics[width = \linewidth]{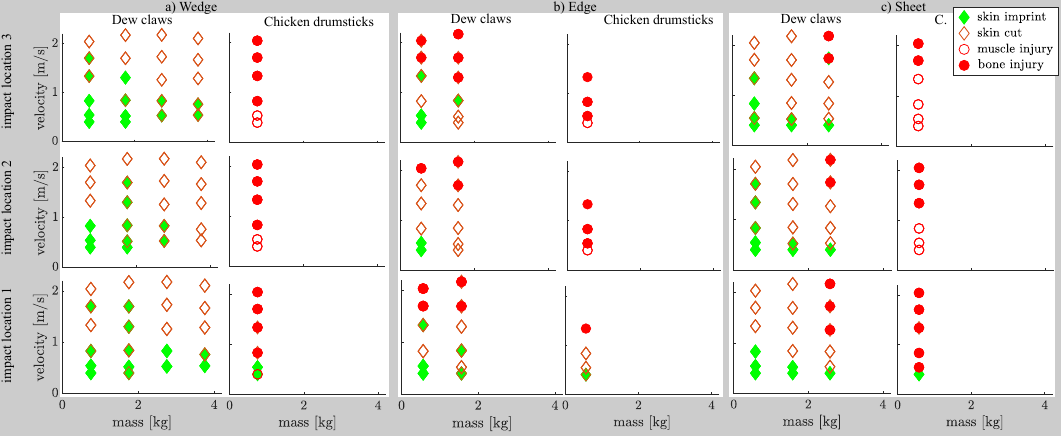}
 \caption{Injury datasets obtained with a) wedge impactor, b) edge impact, and c) sheet impactor for pig dew claws and chicken drumsticks on the three different impact locations.}
 \label{fig:all_results}
\vspace{-5mm}





\end{figure*}

\section{Results} \label{sec:results} 

In this study, a total of 117 pig dew claws and 39 chicken drumsticks were tested in 351 and 117 impact experiments. In the following, the results are presented descriptively. First, all obtained injury protection datasets are listed for the two specimen types, as well as for the different impact locations and impactors. Then, injury probabilities and impact forces corresponding to the injury types are listed for the different impactors and specimens. Finally, validation results are presented. Please note that we do not perform a statistical analysis, as the variety of test conditions and the small number of repetitions under similar conditions ($n=3$) do not allow any statistical conclusions to be drawn. 

\subsection{Injury experiments} 
\begin{figure}[htbp]
\centering
\includegraphics[width=0.9\linewidth]{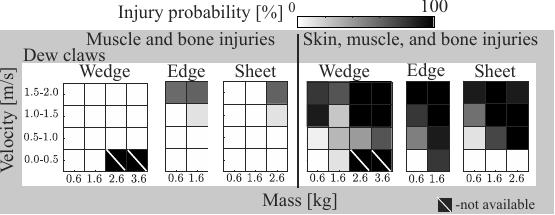}
\caption{Overview of the probability of injuries of pig dew claws grouped by injury types occurring resulting the drop testing.}
\label{fig:prob_injury}
\vspace{-4mm}
\end{figure}

 Fig. \ref{fig:all_results} depicts the three obtained injury prevention datasets for pig dew claw and chicken drumsticks as surrogates for human hands using impactor W, E, and S, subdivided into three impact locations.  
 For the pig dew claw specimens, mainly skin imprints and cuts can be observed for all impactors. In case of impact with the impactor S, bone injuries occurred with velocities higher than \SI{1.0}{m/s} when the mass was \SI{2.6}{kg}. Impactor E caused additionally muscle injury in case of velocities higher than \SI{1.0}{m/s} when the mass was $\geq$ \SI{0.6}{kg}. For the chicken specimen, we observed mainly muscle injuries and bone fractures. Only in the distal impact location on the drumstick skin imprints or cuts occurred. The probability of certain observations for all impact locations clustered by muscle and bone injuries (m/b) and skin cuts, muscle and bone injuries (s/m/b) is provided in Fig. \ref{fig:prob_injury} for the pig specimens. 



\begin{figure}[t]
\centering
\includegraphics[width=0.8\linewidth]{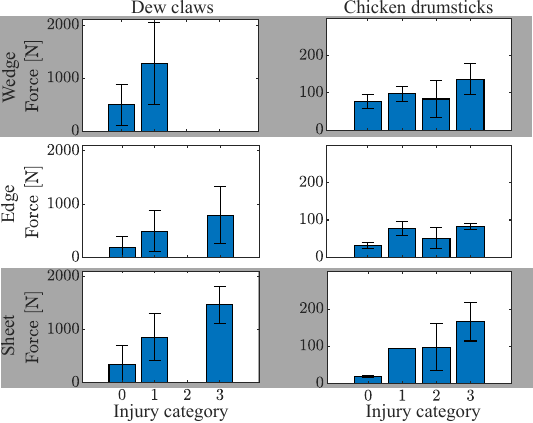}
\caption{Overview of reported forces that caused certain medically observed injuries.}
\label{fig:force_injury}
\vspace{-7mm}
\end{figure}

For descriptive analysis, we link the measured force data to the injury type that occurred during the experiments as depicted in Fig. \ref{fig:force_injury}. For the pig dew claws, skin imprints (0) were observed at \SI{498}{}$\pm$\SI{389}{N}, skin cuts (1) at \SI{1279}{}$\pm$\SI{772}{N}, and no other observations were made using the impactor W. The impactor E caused skin imprints with forces of \SI{197}{}$\pm$\SI{192}{N}, skin cuts occurred with \SI{496}{}$\pm$\SI{391}{N}, and bone injuries (3) were registered at \SI{796}{}$\pm$\SI{531}{N}. For impactor S, skin imprints occurred at \SI{350}{}$\pm$\SI{351}{N}, \SI{860}{}$\pm$\SI{448}{N} caused skin cuts, and bone injuries occurred around \SI{1469}{}$\pm$\SI{346}{N}. The chicken drumsticks showed skin imprints at \SI{77}{}$\pm$\SI{20}{N}, skin cuts at \SI{93}{}$\pm$\SI{27}{N}, muscle injury (2) at \SI{82}{}$\pm$\SI{51}{N}, and bone injuries at \SI{137}{}$\pm$\SI{39}{N} for W. With impactor E, skin imprints occurred at \SI{32}{}$\pm$\SI{8}{N}, the skin was cut at \SI{78}{}$\pm$\SI{18}{N}, muscle injuries occurred at \SI{51}{}$\pm$\SI{28}{N}, and bone injuries at \SI{95}{}$\pm$\SI{22}{N}. With impactor S, the forces were \SI{19}{}$\pm$\SI{3}{N}, \SI{94}{}$\pm$\SI{0}{N}, \SI{98}{}$\pm$\SI{64}{N}, and \SI{168}{}$\pm$\SI{54}{N} for the respective observations.    


\subsection{Validation result}

\begin{table}[tpb]
    \centering
    \caption{Results of the validation studies}
    \vspace{-0.1cm}
    \begin{tabular}{p{10mm}p{3mm}p{3mm}p{2mm}p{4mm}p{10mm}p{3mm}p{3mm}p{2mm}p{4mm}}
        \toprule
        \multicolumn{5}{c}{\textbf{Phillips head}} & \multicolumn{5}{c}{\textbf{Breadboard}}  \\ 
        \midrule
         $v_\mathrm{r}$ [m/s] & $\Sigma_\mathrm{t}$ & $\Sigma_\mathrm{s}$ & $i$ & s/m/b & $v_\mathrm{r}$ [m/s] & $\Sigma_\mathrm{t}$ & $\Sigma_\mathrm{s}$ & $i$ & s/m/b  \\
         \midrule
          0.1 & 19 & 16  & 8 & 0 & 0.2 & 13 & 13 & 13 & 0  \\
        \bottomrule
    \end{tabular}
    \label{tab:val_res}
    \vspace{-0.5cm}
\end{table}

Table \ref{tab:val_res} reports the applied maximum velocity $v_\mathrm{r}$ number of total trials conducted when intending to hit the tied-up pig dew claws on the bone $\Sigma_\mathrm{t}$, the amount of successful contacts with the bone $\Sigma_\mathrm{s}$, the amount of observed skin imprints $i$, and the amount of skin, muscle, or bone injury (s/m/b) observations.




\section{DISCUSSION} \label{sec:Discussion}


 %
We conclude from the generalizable drop test experiments on surrogate specimen and their validation study that by velocity-shaping based on the hand injury protection datasets even in constrained impact scenarios with edged and pointed objects, it can be ensured that collision scenarios with the human hand do not result in open skin, muscle or bone injuries. Impacts with edged or pointed geometries that do not cause such injuries seem feasible with contact velocity $\leq$\SI{0.5}{m/s}, but also with higher velocities up to \SI{1.0}{m/s} depending on the effective robot and payload mass. 
 Thus, of great importance for safely handling pointed or edged objects becomes the robot motion and workspace design considering the effective mass of the system at possible clamping positions. For unconstrained contact situations, the presented datasets can be applied as conservative approximation. Additional research should, nevertheless, consider injury caused by unconstrained contacts as well as shearing contacts.
Overall, we observe strong differences between the two ex-vivo animal surrogates.  Since the bone structure of chickens is hollow in contrast to human hand bones, we observed many scenarios that resulted in bone fractures in the chicken specimen. The pig dew claw bone seems feasible as a human finger bone surrogate based on the observed bone injuries upon contact with the impactor S of the pig claw ($\approx$\SI{1000}{}-\SI{1800}{N}, see Fig. \ref{fig:force_injury}) compared to Hohendorff et al. \cite{Hohendorff201339}, who measured the onset of bone fractures in ex-vivo human fingers clamped in car windows with mostly \SI{1011}{}- \SI{2129}{N} with outliers up to \SI{291}{N}.

Finally, only ex-vivo human testing in the considerable parameter space can validate the use of surrogates and reveal the expectable human injury observations. While ex-vivo specimen experiments allow observation of potential macroscopic skin injuries and similar they do not allow the unambiguous observation of skin contusions (bruises). Only the comparison with in-vivo specimen will allow to differentiate between occurrence of contusions or no injuries after all, e.g., rabbit legs \cite{Fujikawa_2023}. However, the required number of such studies should be minimized as much as possible by prior ex-vivo observations. 

\section{CONCLUSION} \label{sec:Conclusion}

In this work, we propose to create a human hand injury protection database to enable efficient risk assessment and safe integration of pointed or edged tools or objects in pHRI. We start building the database by replacing experiments with generalizable impact studies using a developed drop test device. In this study, we analyze the occurrence of injuries in 351 impacts with pig dew claws and 117 experiments with chicken drumsticks hit by boned tetrahedrons, prisms, and sheet-shaped geometries. In two validation scenarios, we use the obtained datasets to demonstrate their application for safe pHRI design. As a result, we observe that even for constrained impacts with edged or pointed geometries, impacts scenarios can be designed such that no open skin, muscle or bone injuries occur if the effective masses and velocities near pinch points are closely taken into account. This finding opens the door for further investigations and promises to lead to efficient safety implementations for pHRI applications in practice, where edges and corners can never be completely avoided. With a collaborative effort of medical and robotics researchers around the world, we can finally create a complete database of human injuries that takes into account multiple settings and application areas for safe pHRI.

\section*{ACKNOWLEDGMENT}
The authors would like to thank Bach Tran for supporting the test stand design. We gratefully acknowledge the funding of the Lighthouse Initiative KI.FABRIK Bayern by StMWi Bayern, Forschungs- und Entwicklungsprojekt, grant no. DIK0249, the European Union’s Horizon 2020 research and innovation programme as part of the project ReconCycle under grant no. 871352, and of the Bavarian State Ministry for Economic Affairs, Regional Development and Energy (StMWi) as part of the project SafeRoBAY (grant number: DIK0203/01). 
 

\bibliographystyle{IEEEtran} 
\bibliography{References/literature.bib}

\end{document}